\documentclass[journal]{IEEEtran}
\ifCLASSINFOpdf \else \fi

\usepackage{graphicx}
\usepackage{verbatim}
\usepackage[cmex10]{amsmath}
\usepackage{amssymb}
\usepackage{amsfonts}
\usepackage{bbm}

\usepackage{color}

\usepackage{algorithmic}
\usepackage{array}
\usepackage{multirow}
\ifCLASSOPTIONcompsoc
 \usepackage[caption=false,font=normalsize,labelfont=sf,textfont=sf]{subfig}
\else
 \usepackage[caption=false,font=footnotesize]{subfig}
\fi

\usepackage{times}
\usepackage{latexsym}

\usepackage{url}

\graphicspath{ {pic/} }
\usepackage{eucal}
\usepackage[ruled]{algorithm2e}
\usepackage{url}
\usepackage{comment}
\usepackage{footnote}

\ifCLASSINFOpdf
\else
\fi
\begin{document}
\title{Distributed Structured Actor-Critic Reinforcement Learning for Universal Dialogue Management}

% \author{Michael~Shell,~\IEEEmembership{Member,~IEEE,}
%         John~Doe,~\IEEEmembership{Fellow,~OSA,}
%         and~Jane~Doe,~\IEEEmembership{Life~Fellow,~IEEE}% <-this % stops a space
% \thanks{M. Shell was with the Department
% of Electrical and Computer Engineering, Georgia Institute of Technology, Atlanta,
% GA, 30332 USA e-mail: (see http://www.michaelshell.org/contact.html).}% <-this % stops a space
% \thanks{J. Doe and J. Doe are with Anonymous University.}% <-this % stops a space
% \thanks{Manuscript received April 19, 2005; revised August 26, 2015.}}
\author{Zhi Chen, {Lu Chen,~\IEEEmembership{Student Member,~IEEE}, Xiaoyuan Liu, and Kai Yu,~\IEEEmembership{Senior Member,~IEEE}}
\thanks{The work was supported by the National Key Research and Development
Program of China under Grant 2017YFB1002102 and Shanghai Jiao Tong
University Scientific and Technological Innovation Funds (YG2020YQ01). 
 (Corresponding author: Lu Chen and Kai Yu.)

The authors are with the SpeechLab, Department of Computer Science and
Engineering, and MoE Key Lab of Artificial Intelligence, AI Institute, Shanghai
Jiao Tong University, Shanghai 200240, China (e-mail: zhenchi713@sjtu.edu.cn; chenlusz@sjtu.edu.cn; lxy9843@sjtu.edu.cn; kai.yu@sjtu.edu.cn).
}}


% \markboth{IEEE/ACM TRANSACTIONS ON AUDIO, SPEECH, AND LANGUAGE PROCESSING,~Vol.~0, No.~0, December~0}
\markboth{}
{Chen \MakeLowercase{\textit{et al.}}: Distributed Structured Actor-Critic Reinforcement Learning for Universal Dialogue Management}



\maketitle

\begin{abstract}
Traditional dialogue policy needs to be trained independently for each dialogue task.
In this work, we aim to solve a collection of independent dialogue tasks using a unified dialogue agent. The unified policy is parallelly trained using the conversation data from all of the distributed dialogue tasks. However, there are two key challenges:(1) the design of a unified dialogue model to adapt to different dialogue tasks; (2) finding a robust reinforcement learning method to keep the efficiency and the stability of the training process. Here we propose a novel structured actor-critic approach to implement structured deep reinforcement learning (DRL), which not only can learn parallelly from data of different dialogue tasks\footnote{In the experimental setup of this work, each dialogue task has only one dialogue domain.} but also achieves stable and sample-efficient learning. We demonstrate the effectiveness of the proposed approach on 18 tasks of PyDial benchmark. The results show that our method is able to achieve state-of-the-art performance.
\end{abstract}


% \fi
\IEEEpeerreviewmaketitle

\section{Introduction}

\label{sect:introduction}
\IEEEPARstart{T}{he} task-oriented spoken dialogue system (SDS) aims to assist a human user in accomplishing a specific task (e.g., hotel booking). The dialogue management is a core part of SDS. There are two main missions in dialogue management: dialogue belief state tracking (summarising conversation history) and dialogue decision-making (deciding how to reply to the user). In this work, we only focus on devising a policy that chooses which dialogue action to respond to the user.

 The sequential system decision-making process can be abstracted into a partially observable Markov decision process (POMDP)~\cite{young2010hidden}. Under this framework, reinforcement learning approaches can be used for automated policy optimization. 
 In the past few years, there are many deep reinforcement learning (DRL) algorithms， which use neural networks (NN) as function approximators, investigated for dialogue policy ~\cite{su2017sample,peng2018adversarial,lipton2016efficient,chen2017agent,li2017end}.
\begin{figure}%[htbp!]
\centering
\includegraphics[width=0.47\textwidth]{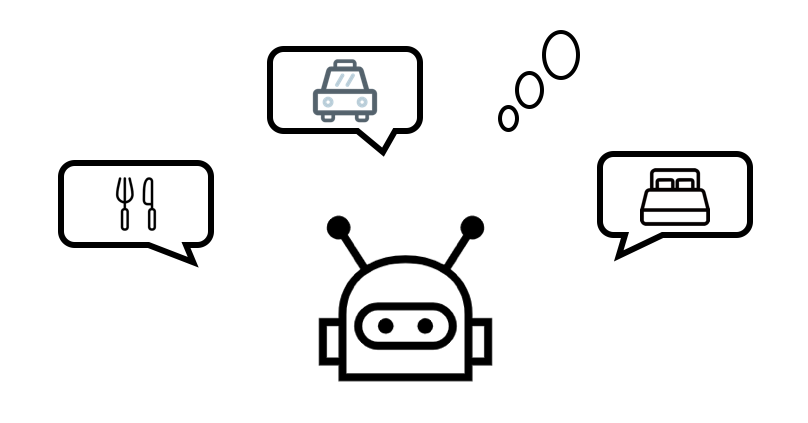}
\caption{The dialogue agent can help different users across different domains parallelly. The generic dialogue agent needs to adapt to different dialogue domains.}
\label{fig:bot}
\end{figure}
Most of these approaches focus on dialogue policy optimization in a single dialogue task. However, in real-life scenarios, a dialogue agent can be asked by many users for different dialogue tasks, e.g., Apple Siri can support many dialogue tasks (find a restaurant, call a taxi, or reserve a hotel as shown in Fig.~\ref{fig:bot}). 
In the multi-task setup, the traditional DRL-based approaches have to train an individual policy for each dialogue task.  It means that each dialogue policy has its independent model parameters, whose scale will increase proportionally with the number of the tasks. One solution is to train a {\em generic} policy for all dialogue tasks \cite{gavsic2015distributed}. However, there are two obstacles to do this with traditional DRL-based approaches.
\begin{itemize}
\item{{\bf Scalability}}: The dialogue state spaces and action sets in two dialogue tasks are usually different, because of their different domain ontologies. Therefore, the model structures have to be different, i.e., the neural networks' parameters cannot be fully shared across tasks. The traditional NN-based policy model does not have good scalability. It is the first obstacle to train a generic policy between different dialogue tasks.
\item{{\bf Efficiency}}:  A {\em stable} and {\em efficient} optimization algorithm is needed to update the parameters of policy using experiences from different dialogue tasks. Most of the traditional DRL algorithms are not sample-efficient, i.e., thousands of dialogues are needed for training an acceptable policy, or the training process is not stable.
\end{itemize}

In this paper, we propose  the \textbf{Str}uctured \textbf{A}ctor-\textbf{C}ritic  Reinforcement Learning for Universal Dialogue Management  ({\bf STRAC}) to address the above two problems. It can use data collected from different dialogue tasks to train a generic policy. To tackle the scalability problem, we utilize the recently proposed {\em structured dialogue policy} \cite{chen2018structured}, where the dialogue policy is represented by a graph neural network (GNN)~\cite{rahimi2018semi, zhou2018graph, wu2019comprehensive}. With the scalability of the GNN (with parameter sharing mechanism and communication mechanism), a single set of parameters can be used for different dialogue tasks. That makes it possible to train a generic policy among multiple dialogue tasks. To tackle the efficiency problem, we deploy an advanced {\em off-policy actor-critic} algorithm, which combines decoupled acting and learning with a novel off-policy correction method called V-trace~\cite{espeholt2018impala}. Combining the improved optimization algorithm with the structured dialogue policy, we can make the generic policy learning process more stable and efficient than the original GNN-based dialogue policy \cite{chen2018structured}. 
We evaluate the performance of STRAC on PyDial benchmark~\cite{ultes2017pydial}, which includes six environments and three dialogue domains. Results show that our unified dialogue agent STRAC gets the best performance on most of the 18 tasks of the benchmark.

\section{Related Work}
\label{sect:relatedwork}
In dialogue management, the multi-domain dialogue tasks can be divided into two types: {\em composite} and {\em distributed}. A composite dialogue task \cite{peng2017composite} consists of serial sub-tasks. The composite dialogue successes only when all the sub-tasks are completed. It is usually formulated by options framework \cite{sutton1998intra} and solved using hierarchical reinforcement learning methods~\cite{budzianowski2017sub,peng2017composite,tang2018subgoal}. The distributed multi-domain task usually tries to find a generic dialogue policy that can be shared by different dialogue domains. In this paper, we pay attention to solve the second multi-domain task. To be clear, we use the multi-task problem to represent the distributed multi-domain problem.

{\bf Distributed Dialogue Policy Optimisation}: The earlier attempts to train a generic dialogue policy for the multi-task problem are based on distributed Gaussian process reinforcement learning (GPRL)~\cite{gavsic2015distributed,wang2015learning}. However, the computational cost of GPRL increases with the increase of the number of data. It is therefore questionable as to whether GPRL can scale to support commercial wide domain SDS \cite{su2017sample, papangelis2019single}. Compared with distributed GP-based methods, STRAC is a distributed NN-based approach with better scalability. Most recently, another NN-based generic policy (DQNDIP) \cite{papangelis2019single} has been proposed, which directly divided the whole dialogue policy by the slots into several sub-policies. These sub-policies have fixed slot-related Domain Independent Parameterisation (DIP) feature~\cite{wang2015learning} as input and pre-defined dialogue actions as output. Because DQNDIP has not modeled the relations among the slots, this approach has an implicit assumption that all the slot-related actions are independent with each other. Compared with DQNDIP, STRAC explicitly considers the relations among the slots through the communication mechanism of the GNN.

{\bf Actor-Critic RL}: 
In recent years, a few actor-critic algorithms are investigated for dialogue policy optimisation, including A2C~\cite{jurvcivcek2011natural,fatemi2016policy}, eNAC~\cite{su2017sample}, and ACER~\cite{weisz2018sample}. Among them, ACER is an efficient off-policy actor-critic method. Unlike traditional actor-critic methods, ACER adopts experience replay and various methods to reduce the
bias and variance of function estimators.
However, it is used in single dialogue tasks, and cannot directly be used to train a generic policy in the multi-task problem.

{\bf Structured Dialogue Policy}: There are two similar structured DRL-based policies with our proposed STRAC. Feudal Dialogue Management (FDM) \cite{casanueva2018feudal,casanueva2018feudal2} directly decomposes the dialogue policy into three kinds of sub-policies. At each turn, a master policy in FDM first decides to take either a slot-independent action or a slot-dependent action. Then the chosen slot-dependent or slot-independent policy is used to choose a primitive action further. Each type of dialogue policy has its private replay memory during the training phase, and its parameters are updated independently. In STRAC, we implicitly decompose a single decision into two-level decisions at each turn, choosing sub-agent first and then selecting the greedy action of the chosen sub-agent. Since there is only one policy in STRAC, the complexity of the training phase does not increase.

Another structured dialogue policy is the recently proposed graph-based policy \cite{chen2018structured}, in which a graph neural network (GNN) is used to coordinate the final decision among all the slot-dependent agents and slot-independent agent. The graph-based dialogue policy is optimized by DQN algorithm using in-domain data. In STRAC, we adopt a more efficient and more stable off-policy actor-critic algorithm to train a generic dialogue policy using all available data collected from different dialogue tasks. 

%\IEEEpubidadjcol

\section{Background}
\label{sect:background}
In slot-filling SDSs, the belief state space $\mathcal{B}$ is defined by the domain ontology, which consists of concepts (or slots) that the dialogue system can talk about. Each slot can take a value from the candidate value set. The user goal can be defined as slot-value pairs, e.g. \emph{\{food=chinese, area=east\}}, which can be used as a constraint to frame a database query. In order to transfer knowledge among domains, the belief state $\mathbf{b}$ can be decomposed into some \emph{slot-dependent} belief states and a \emph{slot-independent} belief state. In order to abstract the state space, slot-independent belief state is extracted from handcrafted feature function $\phi_{0}(\mathbf{b})$ and the $i$-th slot-dependent belief states are extracted from handcrafted feature functions $\phi_{i}(\mathbf{b})$ ($1 \leq i \leq n$) based on the Domain Independent Parameterisation (DIP)~\cite{wang2015learning}, where all the \emph{slot-dependent} belief states have the same dimension. Similarly, the dialogue actions $\mathcal{A}$ can be either slot dependent (e.g. \emph{request}(\emph{food}), \emph{select}(\emph{food}), \emph{confirm}(\emph{food}), \emph{request}(\emph{area}) ,$\cdots$) or slot independent (e.g. \emph{repeat}(), \emph{inform}(),$\cdots$). For each slot, there are three primary actions: \emph{request}($\cdot$), \emph{select}($\cdot$), \emph{confirm}($\cdot$). The number of the slot-dependent primary actions is proportional to the number of the slots. Thus, the whole action space $\mathcal{A}$ can be represented as $\mathcal{A}_{0} \cup \mathcal{A}_{1} \cup \mathcal{A}_{2} \cdots \cup \mathcal{A}_{n}$, where $\mathcal{A}_{i}$ $(1 \leq i \leq n)$ is the set of the $i$-th slot-dependent actions and $\mathcal{A}_{0}$ is the set of slot-independent actions. 

\begin{figure}%[htbp!]
\centering
\includegraphics[width=0.47\textwidth]{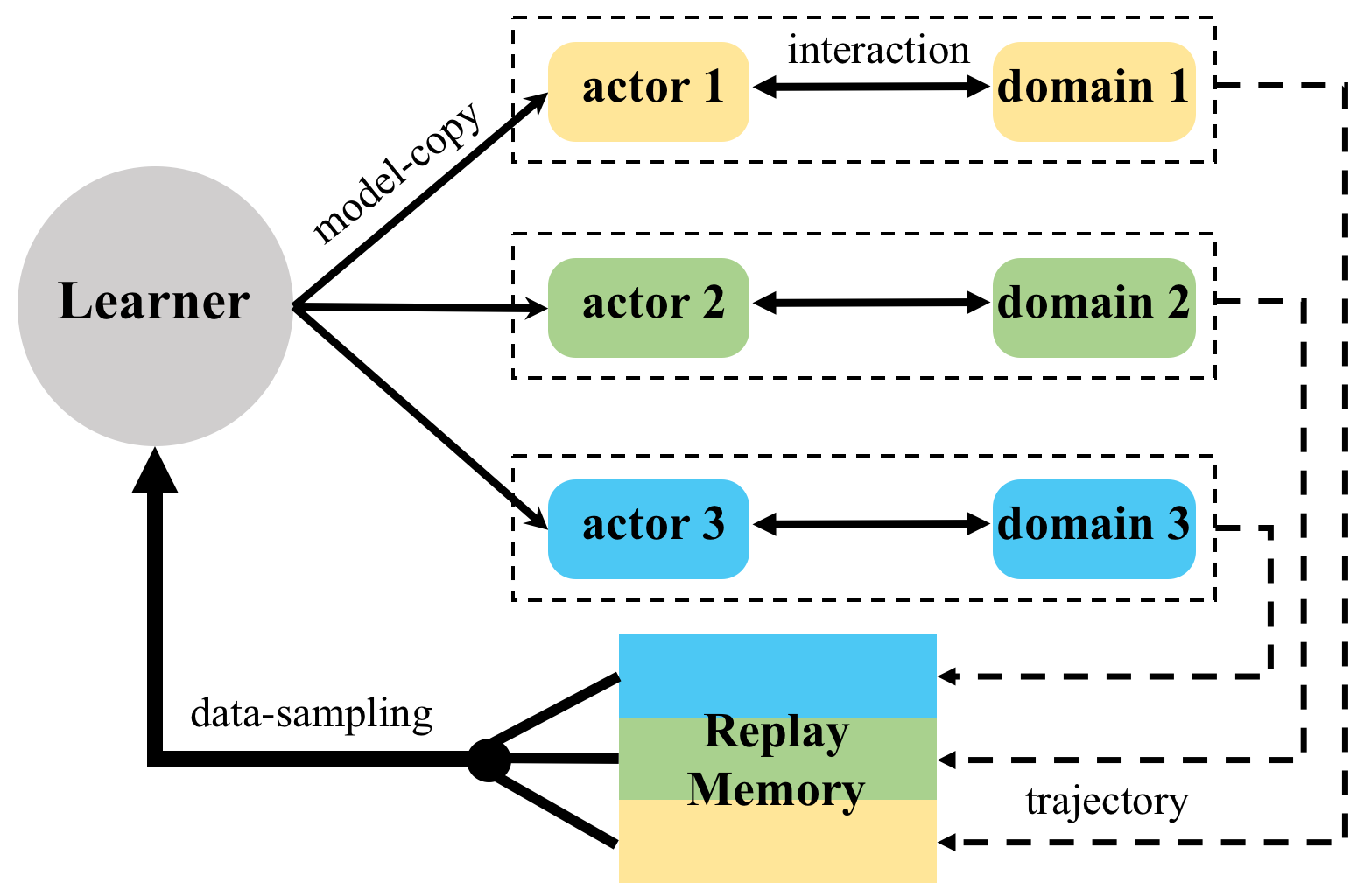}
\caption{The actor-learner architecture for multiple dialogue tasks. The central learner is updated by sampled data from three different dialogue domains.}
\label{fig:LS}
\end{figure}

Dialogue management is a mapping function from belief state space to dialogue action space. The whole response process can be cast as a POMDP with a continuous state space during a dialogue, which can be optimized by reinforcement learning (RL) approaches automatically. The popular RL approaches used to optimize the dialogue policy can be divided into two classes: Q-learning method and policy gradient method. The Q-learning method aims to find the most optimal action-value function by choosing the greedy action, while the policy gradient method tries to find the most optimal policy directly. No matter what kinds of method it is, the objective of RL is to find a policy that maximizes the expected discounted return. In the Q-learning method, the action-value function is optimized by approximating the accumulated greedy action value. Maximizing accumulated return is equivalent to maximizing the value of the initial state $\mathbf{b}_0$, which is estimated by value function. As for the policy gradient method, the optimization of the policy parameters is usually implemented by a stochastic gradient ascent in the direction of $\nabla V^{\pi}(\mathbf{b}_0)$,
\begin{equation}
\nabla_{\theta} V^{\pi}(\mathbf{b}_0) = \mathbb{E}_{\pi}[\sum_{k \geq 0}\gamma^{k}\nabla_{\theta}\pi_{\theta}(a_k|\mathbf{b}_k)Q^{\pi}(\mathbf{b}_k,a_k)],
\label{eq:vgradient}
\end{equation}
where $\gamma$ is a discount factor. $V^{\pi}(\cdot)$ is value function with policy $\pi$. $\theta$ represents the parameters of the policy $\pi$, $\mathbf{b}_k \in \mathcal{B}$ and $a_k \in \mathcal{A}$ are the belief state and the dialogue action at $k$-th dialogue turn respectively. $Q^{\pi}(\mathbf{b}_k,a_k) \overset{\rm def}{=} \mathbb{E}_{\pi}[\sum_{t\geq k}\gamma^{t-k}r_t|\mathbf{b}_k,a_k]$ is the action value of policy $\pi$ at ($\mathbf{b}_k$,$a_k$). In this case, the data used to update the value function are produced under the current policy $\pi$, which means that all the pre-produced data under other policies are unusable. This is an {\em on-policy} method~\cite{sutton2018reinforcement} to estimate state value. It is worth noting that the {\em on-policy} method does not support to reuse the experience data.

Actor-critic method~\cite{konda2000actor} is an important variation of the policy gradient RL method. 
In order to reduce the variance, the advantage function $A_{\beta}$ is usually used in place of the Q-function:
\begin{equation}
\nabla_{\theta} V^{\pi}(\mathbf{b}_0) = \mathbb{E}_{\pi}[\sum_{k \geq 0}\gamma^{k}\nabla_{\theta}\pi_{\theta}(a_k|\mathbf{b}_k)A_{\beta}].
\label{eq:vgradient}
\end{equation}
In actor-critic setting, the advantage function is approximated as $r_k + \gamma v_{k+1} - V_{\beta}(\mathbf{b}_k)$, where $r_k$ is given by reward function, $V_{\beta}(\cdot)$ is output of the state value function, $\beta$ is the parameter of the state value function, $v_{k+1}$ is the estimated state value of the next state which can also be estimated by state value function.
In other words, different from the traditional Policy Gradient method~\cite{sutton2000policy}, the actor-critic method uses another value function to estimate the state value rather than directly calculating from the sampled trajectory data to make the training process more stable. Thus, the actor-critic method has two estimated functions, {\bf policy function} $\pi_{\theta}(\cdot)$ and {\bf state value function} $V_{\beta}(\cdot)$.

\section{STRAC: Structured Actor-Critic for Generic Dialogue Policy}
\label{sect:STRAC}

In this work, we assume that the spoken language understanding module, the belief tracker and the natural language generator can deal with multiple dialogue tasks. We aim to design a unified dialogue agent that can be trained on multiple tasks. 

As shown in Fig.~\ref{fig:LS}, we use an actor-learner architecture to learn the generic policy $\pi$. The agent consists of a set of actors interacting in different dialogue tasks. Each actor interacts with one of these different dialogue tasks and repeatedly generates trajectories of experience saved in the replay memory. 
%There are different experience types collected from different domains. 
In addition to these actors, there is a central learner, which continuously optimizes its policy using the stored trajectories. In this work, we update the central learner's policy once a new trajectory of experience is saved in the replay memory. At the beginning of each dialogue, the corresponding actor updates its own policy $\mu$ to the latest central learner policy $\pi$ and interacts with users for $n$ turns in its dialogue task.

The first challenge we have to overcome is to design a generic policy model that can be trained on all available data collected from different dialogue tasks. Here, our structured generic policy model is improved from ~\cite{chen2018structured} in which Q-learning method~\cite{watkins1992q} is used. Q-learning is a value-based reinforcement learning approach that is easy to be affected by noisy data~\cite{sutton2018reinforcement}. Instead, we adopt a more stable actor-critic method to the structured policy model. This specific model structure will be introduced in Section~\ref{sec:policy}. 

As we know, it is also challenging to keep the training process for a generic policy stable, especially for the actor-critic RL method, which directly returns the distribution of the action space rather than state value. There are two main factors to affect the stability of the optimization process of the generic policy.

\begin{itemize}
\item There is a policy-gap between learner policy $\pi$ and actor policy $\mu$. We use the experiences from policy $\mu$ to update policy $\pi$. Under different policies, the state distribution and the action probability under the same state are both different. They will seriously affect the process of convergence.

\item The trajectories of experience in the replay memory are collected from different dialogue tasks. The state values of different dialogue tasks may differ in scale, which will impact the optimization of the generic policy. The experiences from different tasks will further destabilize the policy updating process.
\end{itemize}

We alleviate the above two problems by deploying a structured DRL approach STRAC, which combines the structured dialogue policy (GNN-based) model with the novel off-policy actor-critic algorithm. The well-designed policy model aims to solve the scalability problem, and the off-policy actor-critic algorithm is used to improve sampling efficiency. In the next two sections, we first introduce the GNN-based dialogue policy with our proposed hierarchical mechanism and then give the structured actor-critic algorithm STRAC.

\begin{figure*}%[htbp!]
\centering
\includegraphics[width=\textwidth]{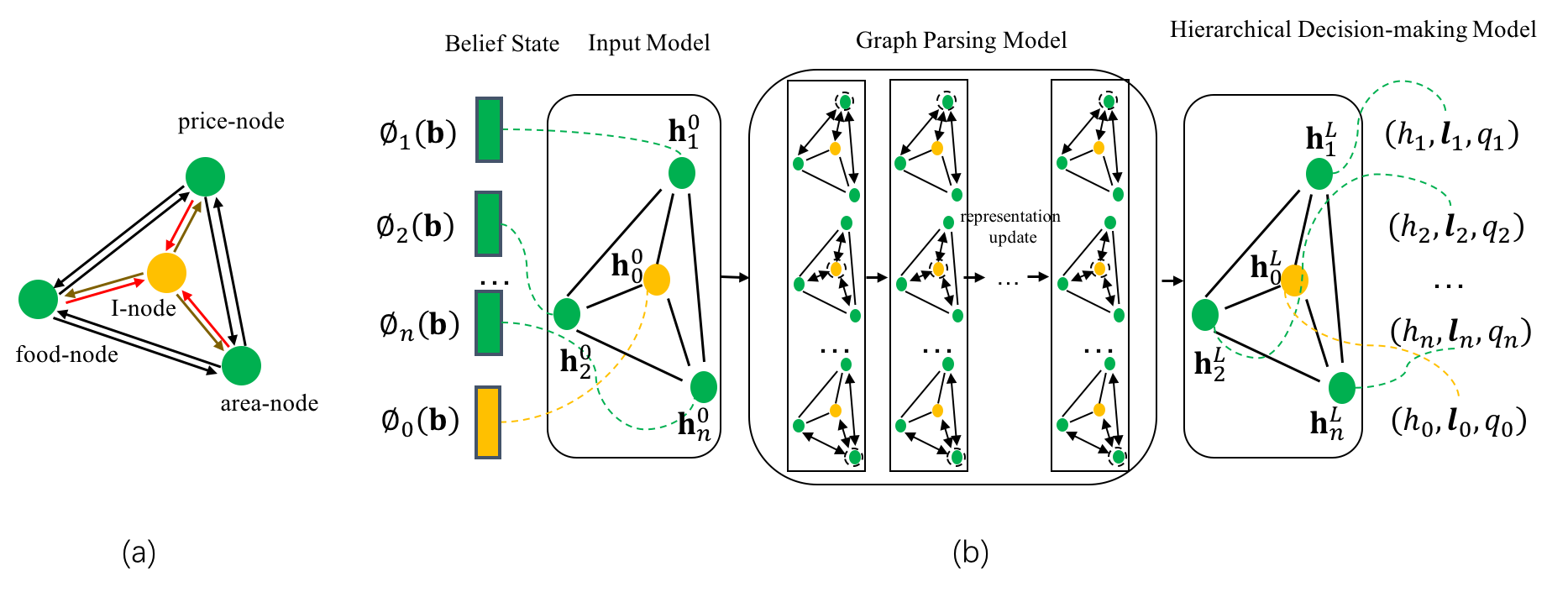}
\caption{a) There are three kinds of S-nodes (price-node, area-node and food-node) in restaurant domain. The I-node represents slot-independent sub-agent. All the sub-agent nodes connect with each other. In this fully-connected graph, there are two kinds of nodes and three kinds of edges. For different dialogue domains, the only difference is the number of the S-node; b) To be easy to interpret, the fully-connected graph shown in (a) is represented by an undirected graph. The structured policy (GNN-based) model consists of three parts: input model, graph parsing model and output model. The initial input of the GNN node is DIP feature of the corresponding slot. The necessary outputs $(h_i, \textbf{l}_i, q_i),0 \leq i \leq n$ of actor-critic approach come from the highest level representation of GNN nodes.}
\label{fig:graph}
\end{figure*}

\subsection{Structured Policy with Hierarchical Mechanism}
\label{sec:policy}
In this section, we introduce how to design a GNN-based policy model with a hierarchical mechanism to represent policy function $\pi_{\theta}(\cdot)$ and state value function $V_{\beta}(\cdot)$ in the actor-critic method. Then, we explain how to solve the scalability problem using the GNN-based policy model, as discussed in Section~\ref{sect:introduction}.

According to the structured decomposition of state space and action space introduced in Section~\ref{sect:background}, the dialogue agent can be divided into two kinds of smaller sub-agents, namely \emph{S-agent} for slot-dependent decision-making and \emph{I-agent} for slot-independent decision-making. The input and output of the $i$-th ($0 \leq i \leq n$) sub-agent are $\phi_{i}(\mathbf{b})$ and probabilities of $\mathcal{A}_{i}$ respectively. The simplest idea assumes that all the sub-agents are independent of each other. In other words, the relations among all the sub-agents are ignored. This kind of strict assumption would decrease performance~\cite{chen2018structured}.  
Instead, the relationships among all the sub-agents can be represented by a graph ~\cite{chen2018structured}, where each node of the graph denotes a corresponding sub-agent, and the directed edge between two sub-agents means that the starting sub-agent will send some information (or message) to the ending sub-agent, as shown in Fig~\ref{fig:graph}(a). The S-agent and I-agent are named by \emph{S-node} and \emph{I-node} respectively in the graph. The input of each sub-agent in the graph is the corresponding belief state. Then, we use GNN to generate the high-level representation of all the nodes from the initial belief state, where all the final node representations contain the graph information. In this paper, we adopt the fully-connected graph to represent their relationships, which means the nodes will communicate with each other. There are three edge types, \texttt{S2S}, \texttt{S2I}, and \texttt{I2S}, which represent the edge type from S-node to another S-node, the edge type from S-node to I-node and the edge type from I-node to S-node respectively. For example, as shown in Fig~\ref{fig:graph}(a), there are three slots: {\em price}, {\em area} and {\em food} in restaurant domain. We calculate the action preference values based on the corresponding high-level node representation. Because each action is corresponding to a sub-agent, the decision-making process can be divided into two levels: {\bf slot level} and {\bf primitive action level}, e.g., the action {\em inform-price} is selected by price agent, where we first find that the action is price related and then choose the primary {\em inform-price} action. In this work, we achieve this hierarchical decision-making mechanism with the advantage of the structured policy model.

We first introduce the GNN-based dialogue policy, which includes three different modules: input module, graph parsing module, and hierarchical decision-making module, as shown in Fig~\ref{fig:graph}(b).

\subsubsection{Input Model}
Before each prediction, each node $v_i$ of the sub-agent graph will receive the corresponding belief state $\phi_i(\mathbf{b})$, which is fed into an input module to obtain a state representation $\mathbf{h}_i^0$ as follows:
\begin{align*}
\mathbf{h}_i^0 = F_{n_i}(\phi_i(\mathbf{b})),
\end{align*}
where $F_{n_i}(\cdot)$ is a function for node type $n_i$ (S-node or I-node), which is a multi-layer perceptron (MLP) in our implementation. According to Section~\ref{sect:background}, the dimensions of all the slot-dependent states are the same. Thus, the input dimension of nodes with the same type is the same. All the S-nodes share the parameters of this model. Thus, there are two input function types: $F_S$ and $F_I $ for S-node and I-node, respectively.

\subsubsection{Graph Parsing Model}
The graph parsing module takes $\mathbf{h}_i^0$ as the initial representation for node $v_i$, then further propagates the higher representation for each node in the graph. The propagation process of node representation at each extract layer is the following.

% , which consists of three steps: message sending, message aggregation and representation update.

{\bf Message Sending} At $l$-th step, for every node $v_i$, there is a node representation $\mathbf{h}_i^{l-1}$. For the other nodes $v_j$ ($i\not= j$), node $v_i$ sends a message vector as below,
\begin{align*}
\mathbf{m}_{ij}^l = M_{c_e}^l(\mathbf{h}_i^{l-1}),
\end{align*}
where $c_e$ is edge type from node $v_i$ to node $v_j$ and $M_{c_e}^l(\cdot)$ is the message generation function which is a linear layer in our implementation: $M_{c_e}^l(\mathbf{h}_i^{l-1}) = \mathbf{W}_{c_e}^l\mathbf{h}_i^{l-1}$. Note that the edges of the same edge type share the trainable weight matrix $\mathbf{W}_{c_e}^l$. In our experiments, there are three types of message computation weights: $\mathbf{W}_{S2S}^l$, $\mathbf{W}_{S2I}^l$ and $\mathbf{W}_{I2S}^l$.

{\bf Message Aggregation} After every node finishes computing message, The messages sent from the other nodes of each node $v_j$ will be aggregated. Specifically, the aggregation process shows as follows:
\begin{align*}
\overline{\mathbf{m}}_{j}^l = A(\{\mathbf{m}_{ij}^l|i\not= j\}),
\end{align*}
where $A(\cdot)$ is the aggregation function which is an average function in our implementation. $\overline{\mathbf{m}}_{j}^l$ is the aggregated message vector which includes the information sent from the other nodes.

{\bf Representation Update} Until now, every node $v_i$ has two kinds of information, the aggregated message vector $\overline{\mathbf{m}}_{i}^l$ and its current representation vector $\mathbf{h}_i^{l-1}$. The representation update process shows as below:
\begin{align*}
\mathbf{h}_i^{l} = U_{n_i}^l(\mathbf{h}_i^{l-1}, \overline{\mathbf{m}}_{i}^l),
\end{align*}
where $U_{n_i}^l(\cdot)$ is the update function for node type $n_i$ at $l$-th parsing layer, which is a non-linear operation in our implementation, 
\begin{align*}
\mathbf{h}_i^{l} = \delta(\mathbf{W}_{n_i}^{l}\mathbf{h}_i^{l-1} + \overline{\mathbf{m}}_{i}^l),
\end{align*}
where $\delta(\cdot)$ is an activation function (e.g., ReLU~\cite{nair2010rectified}), and $\mathbf{W}_{n_i}^{l}$ is a trainable matrix. It is worth noting that the subscript $n_i$ indicates that the nodes of the same node type share the same instance of the update function, here the parameter $\mathbf{W}_{n_i}^{l}$ is shared. There are two types of update weight: $\mathbf{W}_{S}^{l}$ and $\mathbf{W}_{I}^{l}$ for S-node and I-node respectively.

\subsubsection{Hierarchical Decision-making Model}
After updating node representation $L$ steps, each node $v_i$ has the corresponding node representation $\mathbf{h}_i^{L}$. The policy aims to predict a primitive dialogue action. If the dialogue action space is large, it is difficult to choose primitive action directly. Instead, the hierarchical mechanism will be a good choice to simplify the decision-making process, as done by FDM~\cite{casanueva2018feudal,casanueva2018feudal2}.

If we regard slot-independent nodes as a special kind of slot-dependent nodes, each primitive dialogue action can correspond to a slot node. Thus, we can use each node representation $\mathbf{h}_i^{L}$ to calculate a slot-level preference $h_i$ and a primitive preference $\mathbf{l}_i$ for the corresponding primitive actions. The dimension of $\mathbf{l}_i$ equals to the size of $\mathcal{A}_i$. Instead of directly calculating the state value, we calculate a slot-level value $q_i$. The calculation processes of the above three values show as below:
\begin{align*}
h_i = O_{n_i}^{sp}(\mathbf{h}_i^{L}), \ \
\mathbf{l}_i = O_{n_i}^{pp}(\mathbf{h}_i^{L}), \ \
q_i = O_{n_i}^{sq}(\mathbf{h}_i^{L}), 
\end{align*}
where  $O_{n_i}^{sp}(\cdot)$, $O_{n_i}^{pp}(\cdot)$ and $O_{n_i}^{sq}(\cdot)$ are slot preference function, primitive action preference function and slot value function, which are MLPs in practice. Similarly, the subscript $n_i$ indicates that the nodes of the same node type share the same instance of the output functions. Inspired by Dueling DQN~\cite{wang2016dueling}, the preference function of the flat format in each sub-agent can be calculated by:
\begin{equation}
\mathbf{f}_{i} = h_{i} + (\mathbf{l}_{i} - \max(\mathbf{l}_{i})).
\label{eq:pref}
\end{equation}
When predicting a action, all the $\mathbf{f}_i$ will be concatenated, i.e. $\mathbf{f} = \mathbf{f}_1\oplus \dots \oplus \mathbf{f}_n \oplus \mathbf{f}_0$, then the primitive action is chosen according to the distribution $\pi = softmax(\mathbf{f})$. The probability of the slots who the chosen action corresponds to is $\mathbf{p}_{slot} = softmax(\mathbf{h})$, where $\mathbf{h}$ is the concatenation of all $h_i$. 

So far, we may notice that there is no state value function in our structured policy. Here we calculate state value $V$ according to the relation between $Q$-value and $V$-value:
\begin{equation}
V = \mathbf{p}_{slot}^{\top}\cdot \mathbf{q},
\label{eq:q2v}
\end{equation}
where $\mathbf{q}$ is the concatenation of all the slot-level values $q_i$.
Thus, the policy function $\pi$ and the state value function $V$ can be represented by ($q_{i}$,$\mathbf{l}_{i}$,$h_{i}$), $i\in \{0,1,\cdots,n \}$.

\subsubsection{Overview of Decision Procedure}
\label{sec:policyDP}
In this subsection, we explain how to make an implicit hierarchical decision at each turn in STRAC. The high-level preference value of $i$-th slot-node (or sub-agent) is $h_i$. The low-level preference value of the $j$-th dialogue action of $i$-th sub-agent  is $\mathbf{l}_i^j$. According to Equation~\ref{eq:pref}, the final preference value of the $j$-th dialogue action of $i$-th sub-agent is 
\begin{equation}
h_{i} + (\mathbf{l}_{i}^j - \max(\mathbf{l}_{i})).
\label{eq:feq}
\end{equation}
According to the above equation, we know that the final preference value of each dialogue action cannot be greater than the high-level preference of the corresponding sub-agent. In other words, the greatest of the final preference values is equal to the corresponding high-level preference in each sub-agent. When making a decision at each turn, we choose the dialogue action, which has the greatest of the final preference value. Logically, it is equal to a two-level decision-making procedure, first choosing among all sub-agents and then choosing corresponding greedy action in the chosen sub-agent.

When the dialogue action space is vast, this implicit hierarchical decision-making mechanism decomposes the \emph{flat} decision procedure into two simpler decision procedures, which have smaller action spaces. Compared with FDM, STRAC is a \emph{differentiable} end-to-end hierarchical framework in the actor-critic setting.

On the other hand, assume that the dialogue task has changed. For the GNN policy, the slot-dependent nodes would increase or decrease. However, in the GNN policy, the slot-dependent nodes share the parameters. Thus, the parameters in the GNN policy are fixed for different dialogue tasks. This property of the GNN policy is the fundamental of STRAC. In Section~\ref{sec:IMPALA}, we will introduce the training procedure of the structured actor-critic algorithm.

\subsection{Policy Training}
\label{sec:IMPALA}
Until now, the scalability problem has been solved by using a GNN-based policy model with a hierarchical mechanism to represent policy function $\pi$ and state value function $V$ in the actor-critic approach. 
In this subsection, we will introduce how to update the parameters of $\pi$ and $V$ using the off-policy actor-critic method and explain how to learn a generic policy for the multi-task problem.

% As introduced in Section~\ref{sect:background}, the gradient of the policy in the off-policy actor-critic algorithm is calculated by Equation~\ref{eq:vgradient2}. 

Different from  the on-policy methods, the off-policy methods try to reuse the previous experience data which is normally stored in experience memory to improve the sampling efficiency. For the Q-learning methods, the decision-making is according to the action value function. Because the reward function of a dialogue task is normally fixed, the reward value stored in the experience data does not change with Q-learning policy. Thus, the experience data can be directly used to update the current Q-learning policy (action value function). Different from Q-learning method, the policy gradient (PG) method directly estimates the action probability. For the same trajectory data, two different PG policies usually have two different probabilities to produce this trajectory. Therefore, there is a gap to estimate the value of this trajectory between these two different policies. This is why the previous experience data produced by another policy can not be used to update the current policy directly for the policy gradient RL methods. According to the policy gradient theorem~\cite{sutton2000policy, kakade2002natural} and importance sampling theorem~\cite{liu1996metropolized, neal2001annealed}, the policy approximation of the true gradient $\nabla_{\theta} V^{\pi}(\mathbf{b}_0)$ in the off-policy policy gradient RL method is:
\begin{align}
&\nabla_{\theta} V^{\pi}(\mathbf{b}_0) = \mathbb{E}_{\pi}[\sum_{k \geq 0}\gamma^{k}\nabla_{\theta}\pi_{\theta}(a_k|\mathbf{b}_k)A_{\beta}(\mathbf{b}_k)] \notag \\
&\approx \mathbb{E}[\sum_{a \in \mathcal{A}}\nabla_{\theta} \pi_{\theta}(a|\mathbf{b})A_{\beta}(\mathbf{b})|\mathbf{b}\sim d^{\mu}] \notag \\
&= \mathbb{E}[\sum_{a \in \mathcal{A}}\mu(a|\mathbf{b})\frac{\pi(a|\mathbf{b})}{\mu(a|\mathbf{b})}\frac{\nabla_{\theta} \pi_{\theta}(a|\mathbf{b})}{\pi_{\theta}(a|\mathbf{b})}A_{\beta}(\mathbf{b})|\mathbf{b}\sim d^{\mu}] \notag \\
&= \mathbb{E}[\sum_{a \in \mathcal{A}} \rho \nabla_{\theta}\log \pi_{\theta}(a|\mathbf{b})A_{\beta}(\mathbf{b})|\mathbf{b}\sim d^{\mu},a\sim \mu(\cdot|\mathbf{b})]. 
\label{eq:vgradient2}
\end{align}
where $\rho = \frac{\pi(a|\mathbf{b})}{\mu(a|\mathbf{b})}$ and $(\mathbf{b},a)$ is generated under the policy $\mu$ and $d^{\mu}$ is the distribution of the belief state under the policy $\mu$.

As discussed in Section~\ref{sect:background}, the advantage function is approximated as $r_k + \gamma v_{k+1} - V_{\beta}(\mathbf{b}_k)$, where $r_k$ is given by reward function, $V_{\beta}(\mathbf{b}_k)$ is output of the state value function according to Equation~\ref{eq:q2v}, $v_{k+1}$ is the estimated state value of the next state. In off-policy setting, because there is a policy gap between the actor policy and learner policy, $v_{k+1}$ can not be directly estimated by the cumulative reward value like the Q-learning method does. Instead, in actor-critic setup, $v_{k+1}$ is calculated by V-trace algorithm~\cite{espeholt2018impala} in order to eliminate the policy-gap.
As demonstrated in ~\cite{espeholt2018impala}, V-trace is a stable off-policy method to estimate the target of the state value, when there is a gap between the actor's policy and the learner's policy. The generic policy is optimized by sampled data from different dialogue tasks. Thus, the sample efficiency problem has also been solved.

%Lastly, we analyze the implicit hierarchical decision-making mechanism in detail.

\begin{algorithm}
\label{alg:strac}
\caption{Distributed Structured Actor-Critic Algorithm}
\LinesNumbered
\KwIn{$D$(domain), $M$(memory), $q$(action value), $\pi$(action-level policy),$\pi_h$(slot-level policy), $n$(steps),$\phi$(state handcraft function), $\beta$(the parameter of value function), $\theta$(the parameter of policy function), hyperparameters:($\overline{c}$, $\overline{\rho}$,$\lambda_1$, $\lambda_2$,$\alpha$)}
Initialise $d_{\theta} = 0$, $d_{\beta} = 0$\;
\For{$\rm each$ $\rm domain$ $\rm d$ $\rm in$ $\rm D$}{
　　\For{$\rm each$ $\rm dialogue$ $\{b_{1:N}, a_{1:N}, r_{1:N}, \mu_{1:N}\}$ $\rm in$ $\rm M$ $\rm sampled$ $\rm from$ $\rm d$}{
　　    $\rho_k \longleftarrow \min (\overline{\rho}, \frac{\pi_\theta(a_k|\mathbf{b}_k)}{\mu_k})$ \;
　　    $c_k \longleftarrow \min ( \overline{c}, \frac{\pi_\theta(a_k|\mathbf{b}_k)}{\mu_k})$\;
　　    $V_{\beta}(\mathbf{b}_k) \longleftarrow \sum_i               \pi_h(\mathcal{A}_i|\phi_{i}(\mathbf{b}_k))q_{i}(\phi_{i}(\mathbf{b}_k))$\;
　　    $v_k  \longleftarrow  V_{\beta}(\mathbf{b}_k) + \sum_{s=k}^{s=k+n-1}\gamma^{s-k}(\prod_{d=k}^{s-1}c_d)(\rho_s(r_s+\gamma V_{\beta}(\mathbf{b}_{s+1})-V_{\beta}(\mathbf{b}_s)))$\;
　　    $\nabla_{\beta} \longleftarrow (v_k-V_{\beta}(\mathbf{b}_k))\nabla_{\beta}V_{\beta}(\mathbf{b}_k)$\;
　　    $A_k \longleftarrow r_k + \gamma v_{k+1} - V_{\beta}(\mathbf{b}_{k})$\;
　　    $\nabla_{\theta} \longleftarrow \lambda_{1}\rho_k \nabla_{\theta} \log \pi_{\theta}(a_k|\mathbf{b}_k)A_k - \lambda_{2} \nabla_{\theta} \sum_a \pi_{\theta}(a|\mathbf{b}_k)\log \pi_{\theta}(a|\mathbf{b}_k)$\;
　　    $d_{\beta} \longleftarrow d_{\beta}-\nabla_{\beta}$\;
　　    $d_{\theta} \longleftarrow d_{\theta}-\nabla_{\theta}$\;
　　}
　　$\theta \longleftarrow \theta+\alpha d_{\theta}/N$\;
　　$\beta \longleftarrow \beta+\alpha d_{\beta}/N$\;
}
\end{algorithm}

\subsubsection{V-trace}
Consider a dialogue trajectory $(\mathbf{b}_t, a_t, r_t)_{t=k}^{t=k+n}$ generated by the actor following some policy $\mu$. According to the V-trace theory, the $n$-step target state value $v_k$ of the state $\mathbf{b}_k$ is defined as:
\begin{equation}
v_k \overset{\rm def}{=} V_{\beta}(\mathbf{b}_k) + \sum_{t=k}^{k+n-1}\gamma^{t-k}(\prod_{d=k}^{t-1}c_d)\delta_k^V, 
\label{eq:vtarget}
\end{equation}
where $V_{\beta}(\mathbf{b}_k)$ is the state function defined in Equation \ref{eq:q2v}, and $\delta_k^V \overset{\rm def}{=} \rho_k(r_k+\gamma V_{\beta}(\mathbf{b}_{k+1})-V_{\beta}(\mathbf{b}_k))$ is a temporal difference for $V$, and $\rho_k \overset{\rm def}{=} \min(\overline{\rho},\frac{\pi(a_k|\mathbf{b}_k)}{\mu(a_k|\mathbf{b}_k)})$ and $c_d \overset{\rm def}{=} \min(\overline{c}, \frac{\pi(a_d|\mathbf{b}_d)}{\mu (a_d|\mathbf{b}_d)})$ are truncated importance sampling weights~\cite{ionides2008truncated}. The weight $\rho_k$ defines the fixed point of this update rule. In other words, the weight $\rho_k$ keeps the convergence of the online V-trace algorithm, proved by \cite{espeholt2018impala}, which is used to correct the instant effects on the current timestep. The weights $c_d$ are used to correct the accumulated effects of previous timesteps (or the trace coefficients in Retrace~\cite{precup2000eligibility}). 
Their product $\prod_{d=k}^{t-1}c_d$ measures how much a temporal difference $\delta_k^V$ observed at time $t$ impacts the update of the value function at a previous time $k$ under the policy $\pi$. The truncation levels $\overline{c}$ and $\overline{\rho}$ have different effects of the V-trace. $\overline{c}$ controls the speed of the convergence to this function. $\overline{\rho}$ impacts the natural level of the convergence. According to Equation~\ref{eq:vtarget}, we can calculate the target of the state value under the policy $\pi$ using collected data in the replay memory. But the cost is that we have to store an additional action probability $\mu(a|\mathbf{b})$ at each timestep in the replay memory.

\subsubsection{Training Procedure for Multi-task Problem}

During the training phrase, the parameters of both state value function $V_{\beta}(\cdot)$ and policy function $\pi_{\theta}(\cdot)$ are updated. The optimisation object of $V_{\beta}(\mathbf{b}_k)$ is to approximate the $n$-step target state value $v_k$, i.e. the loss is the mean square error (MSE) between $v_k$ and $V_{\beta}(\mathbf{b}_k)$, $(v_k-V_{\beta}(\mathbf{b}_k))^2$. Therefore, the  parameters $\beta$ are updated by gradient descent in direction of 
\begin{equation}
\nabla_{\beta} = (v_k-V_{\beta}(\mathbf{b}_k))\nabla_{\beta}V_{\beta}(\mathbf{b}_k).
\label{eq:vloss}
\end{equation}

\begin{center}
\begin{table*}[h]%!hbp
\small
\centering
\caption{Reward and success rates after 400/4000 training dialogues. The results in bold blue are the best success rates, the results in bold black are the best rewards not including Feudal ACER. We also tried to get the learning curves of Feudal ACER, but we cannot get similar results provided in the original paper. We find that Feudal ACER is very susceptible to the seeds especially in Env.2 and Env.4 without the masking mechanism.}
% \footnotetext{We also tried to get the learning curves of Feudal ACER, but we cannot get similar results provided in the original paper. The results of Feudal ACER is directly taken from the original paper. We find that Feudal ACER is very susceptible to the seeds especially in Env.2 and Env.4 without the masking mechanism.}
\begin{tabular}{rccccccccccccccccccccccl}
\hline
 & & \multicolumn{12}{c}{Baselines} & \multicolumn{4}{c}{STRAC} \\
% & & \multicolumn{6}{c}{\fbox{ \ GP-Sarsa \ \qquad DQN \qquad \qquad eNAC \quad}} & \multicolumn{2}{c}{GNN-M} & \multicolumn{2}{c}{GNN-A} & \multicolumn{2}{c}{GNN-M-C} & \multicolumn{2}{c}{GNN-A-C}\\
\hline
& & \multicolumn{2}{c}{GP-Sarsa} & \multicolumn{2}{c}{DQN} & \multicolumn{2}{c}{ACER} & \multicolumn{2}{c}{FeudalDQN}  & \multicolumn{2}{c}{FMGNN} & \multicolumn{2}{c}{DQNDIP-M} & \multicolumn{2}{c}{STRAC-S} & \multicolumn{2}{c}{STRAC-M} \\
 \multicolumn{2}{c}{\emph{Task}} & Suc. & Rew. & Suc. & Rew. & Suc. & Rew. & Suc. & Rew.& Suc. & Rew.& Suc. & Rew.& Suc. & Rew.& Suc. & Rew.\\
 \hline
 \multicolumn{16}{c}{ after 400 training dialogues}  \\
 \hline
 \multirow{3}{*}{\rotatebox{90}{Env.1}}
& CR & 95.6 & 12.7 & 92.3 & 12.3 & 92.7 & 12.4 & 86.8 & 11.2 & 85.9 & 11.0 & 86.2 & 10.9 & 97.7 & 13.1 & {\color{blue} \textbf{99.7}} & \textbf{14.0}\\
& SFR & 91.9 & 10.7 & 58.5 & 4.0 & 67.3 & 6.5 & 34.7 & -0.0 & 52.3 & 3.4 & 93.9 & 11.6 & 98.2 & 12.3 & {\color{blue} \textbf{99.2}} & \textbf{12.9}\\
& LAP & 88.8 & 9.0 & 44.3 & 0.9 & 59.0 & 5.2 & 43.1 & 1.6  & 55.4 & 4.6 & 93.5 & 11.1 & 98.5 & \textbf{12.3} & {\color{blue} \textbf{98.6}} & 12.2\\
 \hline
 \multirow{3}{*}{\rotatebox{90}{Env.2}}
& CR & {\color{blue} \textbf{90.6}} & \textbf{10.9} & 78.1 & 8.6 & 70.4 & 5.6 & 68.1 & 6.0 & 75.5 & 8.1 & 73.7 & 7.0 & 65.5 & 5.0 & 90.3 & 10.2\\
& SFR & 81.7 & 7.7 & 63.6 & 5.0 & 56.9 & 3.4 & 64.1 & 4.8 & 71.2 & 7.1 & {\color{blue} \textbf{92.2}} & \textbf{11.1} & 69.8 & 4.4 & 87.5 & 9.0\\
& LAP & 68.7 & 4.7 & 36.4 & -3.2 & 18.6 & -6.1 & 64.1 & 4.8 & 76.5 & 7.7 & {\color{blue} \textbf{92.2}} & \textbf{10.6} & 56.9 & 1.6 & 89.2 & 9.1\\
 \hline
 \multirow{3}{*}{\rotatebox{90}{Env.3}}
& CR & 84.0 & 9.3 & 79.6 & 8.7 & 85.4 & 10.2 & 90.8 & 11.3 & 78.3 & 8.8 & 92.0 & 11.2 & 97.2 & 12.5 & {\color{blue} \textbf{97.3}} & \textbf{12.7}\\
& SFR & 78.7 & 6.5 & 44.8 & 0.5 & 44.9 & 1.5 & 68.2 & 5.4 & 54.6 & 3.1 & 89.6 & 9.5 & 90.4 & 8.9 & {\color{blue} \textbf{93.6}} & \textbf{10.5}\\
& LAP & 67.0 & 4.1 & 39.6 & -0.5 & 49.8 & 2.8 & 66.5 & 4.8 & 49.6 & 2.5 & 89.9 & 8.9 & {\color{blue} \textbf{92.5}} & \textbf{9.7} & 92.4 & 9.6\\
 \hline
 \multirow{3}{*}{\rotatebox{90}{Env.4}}
& CR & 79.9 & 7.3 & 67.1 & 5.0 & 67.4 & 5.0 & 54.5 & 0.8 & 70.5 & 6.6 & {\color{blue} \textbf{90.4}} & \textbf{10.2} & 71.0 & 5.2 & 75.3 & 6.6\\
& SFR & 73.2 & 5.3 & 55.7 & 1.4 & 33.5 & -1.7 & 50.0 & 1.8 & 66.7 & 5.6 & {\color{blue} \textbf{86.5}} & \textbf{8.9} & 72.7 & 5.0 & 77.2 & 6.4\\
& LAP & 51.7 & 0.1 & 27.0 & -5.1 & 20.7 & -3.8 & 43.9 & -0.5 & 63.7 & 4.6 & {\color{blue} \textbf{85.7}} & \textbf{8.1} & 65.9 & 3.1 & 79.8 & 6.9\\
 \hline
 \multirow{3}{*}{\rotatebox{90}{Env.5}}
& CR & 71.2 & 5.9 & 56.4 & 2.4 & 68.0 & 5.8 & 77.9 & 7.6 & 53.8 & 2.8 & 89.4 & 9.4 & 95.3 & 10.6 & {\color{blue} \textbf{95.6}} & \textbf{10.8}\\
& SFR & 58.2 & 1.2 & 28.5 & -3.8 & 29.3 & -2.3 & 54.7 & 1.3 & 26.0 & -2.9 & 77.3 & 4.9 & 80.6 & 4.5 & {\color{blue} \textbf{88.8}} & \textbf{7.5}\\
& LAP & 30.4 & -2.9 & 22.1 & -5.0 & 28.3 & -2.3 & 51.9 & 0.9 & 23.4 & -3.0 & 80.4 & 4.7 & {\color{blue} \textbf{87.8}} & \textbf{6.1} & 86.0 & 5.6\\
 \hline
 \multirow{3}{*}{\rotatebox{90}{Env.6}}
& CR & 82.7 & 8.1 & 67.3 & 5.4 & 72.8 & 6.9 & 82.1 & 8.7 & 62.8 & 5.3 & 88.0 & 9.5 & {\color{blue} \textbf{91.9}} & \textbf{10.3} & 90.7 & 9.9\\
& SFR & 55.7 & 0.9 & 33.0 & -2.6 & 40.3 & -0.1 & 59.2 & 2.5 & 36.2 & -0.8 & 78.3 & 5.7 & 78.5 & 4.9 & {\color{blue} \textbf{83.8}} & \textbf{6.6}\\
& LAP & 49.2 & -0.6 & 37.3 & -1.5 & 44.1 & 0.6 & 56.3 & 2.2 & 37.2 & -0.3 & 78.5 & 5.3 & {\color{blue} \textbf{84.6}} & \textbf{6.6} & 81.7 & 5.7\\
 \hline
 \multirow{4}{*}{\rotatebox{90}{Mean}}
& CR & 84.0 & 9.0 & 73.5 & 7.1 & 76.1 & 7.6 & 76.7 & 7.6 & 71.1 & 7.1 & 86.6 & 9.7 & 86.4 & 9.5 & {\color{blue} \textbf{91.5}} & \textbf{10.7}\\
& SFR & 73.2 & 5.4 & 47.3 & 0.8 & 45.4 & 1.2 & 55.2 & 2.6 & 51.2 & 2.6 & 86.3 & 8.6 & 81.7 & 6.7 & {\color{blue} \textbf{88.3}} & \textbf{8.8}\\
& LAP & 59.3 & 2.4 & 34.4 & -2.4 & 36.8 & -0.6 & 54.3 & 2.3 & 51.0 & 2.7 & 86.7 & 8.1 & 81.0 & 6.6 & {\color{blue} \textbf{88.0}} & \textbf{8.2}\\
 \hline
 
 \hline
\multicolumn{16}{c}{ after 4000 training dialogues}  \\
 \hline
 \multirow{3}{*}{\rotatebox{90}{Env.1}}
& CR & 98.9 & 13.6 & 91.4 & 12.1 & 99.4 & 13.9 & 78.2 & 9.2 & 74.8 & 8.7 & 96.2 & 13.2 & {\color{blue} \textbf{99.8}} & {\color{black} \textbf{14.1}} & {\color{blue} \textbf{99.8}} & {\color{black} \textbf{14.1}}\\
& SFR & 96.7 & 12.0 & 84.5 & 9.5 & 94.5 & 11.8 & 34.6 & -0.6 & 61.3 & 5.3 & 97.0 & 12.4 & {\color{blue} \textbf{98.7}} & {\color{black} \textbf{12.7}} & 98.5 & {\color{black} \textbf{12.7}}\\
& LAP & 96.1 & 11.0 & 81.8 & 8.5 & 87.9 & 10.0 & 62.5 & 5.2 & 78.5 & 8.6 & 94.2 & 11.4 & 97.6 & {\color{black} \textbf{12.0}} & {\color{blue} \textbf{97.8}} & {\color{black} \textbf{12.0}}\\
 \hline
 \multirow{3}{*}{\rotatebox{90}{Env.2}}
& CR & 97.9 & 12.8 & 88.7 & 11.0 & 87.4 & 11.3 & 90.8 & 10.5 & 93.6 & 12.2 & {\color{blue} \textbf{99.2}} & {\color{black} \textbf{13.9}} & 97.9 & 13.1 & 98.4 & 13.1\\
& SFR & 94.7 & 11.1 & 76.6 & 7.5 & 80.6 & 8.5 & 89.8 & 10.3 & 93.0 & 11.5 & {\color{blue} \textbf{98.0}} & {\color{black} \textbf{13.2}} & 95.6 & 12.1 & 97.5 & 13.0\\
& LAP & 89.1 & 9.9 & 52.0 & 2.2 & 78.0 & 7.7 & 96.0 & 12.1 & 91.4 & 11.1 & 97.9 & 12.7 & 92.6 & 11.6 & {\color{blue} \textbf{98.0}} & {\color{black} \textbf{12.8}}\\
 \hline
 \multirow{3}{*}{\rotatebox{90}{Env.3}}
& CR & 92.1 & 11.1 & 92.1 & 11.5 & 93.7 & 11.7 & {\color{blue} \textbf{98.4}} & {\color{black} \textbf{13.0}} & 96.6 & 12.6 & 96.7 & 12.6 & 98.1 & {\color{black} \textbf{13.0}} & 97.9 & 12.9\\
& SFR & 87.5 & 8.6 & 68.6 & 5.0 & 82.1 & 7.7 & 92.5 & 10.2 & 89.4 & 9.4 & 91.9 & 10.1 & 91.9 & 10.5 & {\color{blue} \textbf{93.0}} & {\color{black} \textbf{10.6}}\\
& LAP & 81.6 & 7.2 & 64.4 & 4.1 & 76.8 & 6.6 & 87.4 & 8.9 & 84.2 & 7.9 & 91.5 & 9.7 & 90.7 & 9.7 & {\color{blue} \textbf{92.1}} & {\color{black} \textbf{9.9}}\\
 \hline
 \multirow{3}{*}{\rotatebox{90}{Env.4}}
& CR & 93.4 & 10.2 & 88.0 & 9.3 & 87.2 & 9.7 & 95.5 & 12.3 & 90.9 & 11.0 & {\color{blue} \textbf{96.4}} & {\color{black} \textbf{12.5}} & 92.9 & 11.5 & 91.3 & 10.8\\
& SFR & 85.9 & 8.6 & 60.3 & 2.7 & 69.6 & 5.8 & 92.9 & 10.8 & 87.7 & 9.6 & {\color{blue} \textbf{94.7}} & {\color{black} \textbf{11.6}} & 90.2 & 10.7 & 89.2 & 10.5\\
& LAP & 73.8 & 5.8 & 53.4 & 0.8 & 61.8 & 4.7 & 94.2 & 11.3 & 83.3 & 8.2 & {\color{blue} \textbf{94.9}} & {\color{black} \textbf{11.4}} & 86.3 & 9.2 & 89.7 & 10.4\\
 \hline
 \multirow{3}{*}{\rotatebox{90}{Env.5}}
& CR & 79.2 & 7.4 & 86.4 & 8.9 & 91.6 & 10.2 & 95.2 & 11.3 & 95.2 & 11.2 & 95.5 & 11.2 & {\color{blue} \textbf{97.1}} & {\color{black} \textbf{11.8}} & 96.5 & 11.7\\
& SFR & 75.9 & 5.2 & 63.5 & 2.2 & 76.0 & 4.8 & 86.7 & 7.5 & 82.3 & 5.9 & 88.4 & 7.9 & 89.6 & {\color{black} \textbf{8.4}} & {\color{blue} \textbf{90.1}} & {\color{black} \textbf{8.4}}\\
& LAP & 46.5 & -0.2 & 50.0 & -0.2 & 61.0 & 1.2 & 80.7 & 5.5 & 70.0 & 2.8 & 84.6 & 6.2 & 88.2 & 6.9 & {\color{blue} \textbf{88.5}} & {\color{black} \textbf{7.0}}\\
 \hline
 \multirow{3}{*}{\rotatebox{90}{Env.6}}
& CR & 89.4 & 9.8 & 85.9 & 9.4 & 86.3 & 8.9 & 89.9 & 10.3 & 89.3 & 10.0 & 90.6 & 10.3 & {\color{blue} \textbf{92.5}} & {\color{black} \textbf{11.0}} & 91.5 & 10.7\\
& SFR & 71.0 & 4.2 & 52.5 & 0.7 & 64.5 & 2.6 & 80.8 & 6.9 & 70.8 & 4.3 & 82.6 & 7.1 & 81.6 & 7.0 & {\color{blue} \textbf{84.5}} & {\color{black} \textbf{7.3}}\\
& LAP & 54.2 & 1.4 & 48.4 & 0.4 & 61.0 & 2.0 & 78.8 & 6.0 & 68.7 & 4.0 & 81.6 & 6.6 & {\color{blue} \textbf{83.3}} & {\color{black} \textbf{6.7}} & 83.2 & {\color{black} \textbf{6.7}}\\
 \hline
 \multirow{4}{*}{\rotatebox{90}{Mean}}
& CR & 91.8 & 10.8 & 88.8 & 10.4 & 90.9 & 11.0 & 91.3 & 11.1 & 90.1 & 11.0 & 95.8 & 12.3 & {\color{blue} \textbf{96.4}} & \textbf{12.4} & 95.9 & 12.2\\
& SFR & 85.3 & 8.3 & 67.7 & 4.6 & 77.9 & 6.9 & 79.6 & 7.5 & 80.8 & 7.7 & {\color{blue} \textbf{92.1}} & \textbf{10.4} & 91.3 & 10.2 & {\color{blue} \textbf{92.1}} & \textbf{10.4}\\
& LAP & 73.6 & 5.8 & 58.3 & 2.6 & 71.1 & 5.4 & 83.3 & 8.2 & 79.4 & 7.1 & 90.8 & 9.7 & 89.8 & 9.4 & {\color{blue} \textbf{91.6}} & \textbf{9.8}\\
 \hline
\end{tabular}

\label{tab:results}
\end{table*}
\end{center}

\begin{figure*}[htbp!]
\centering
\includegraphics[width=\textwidth]{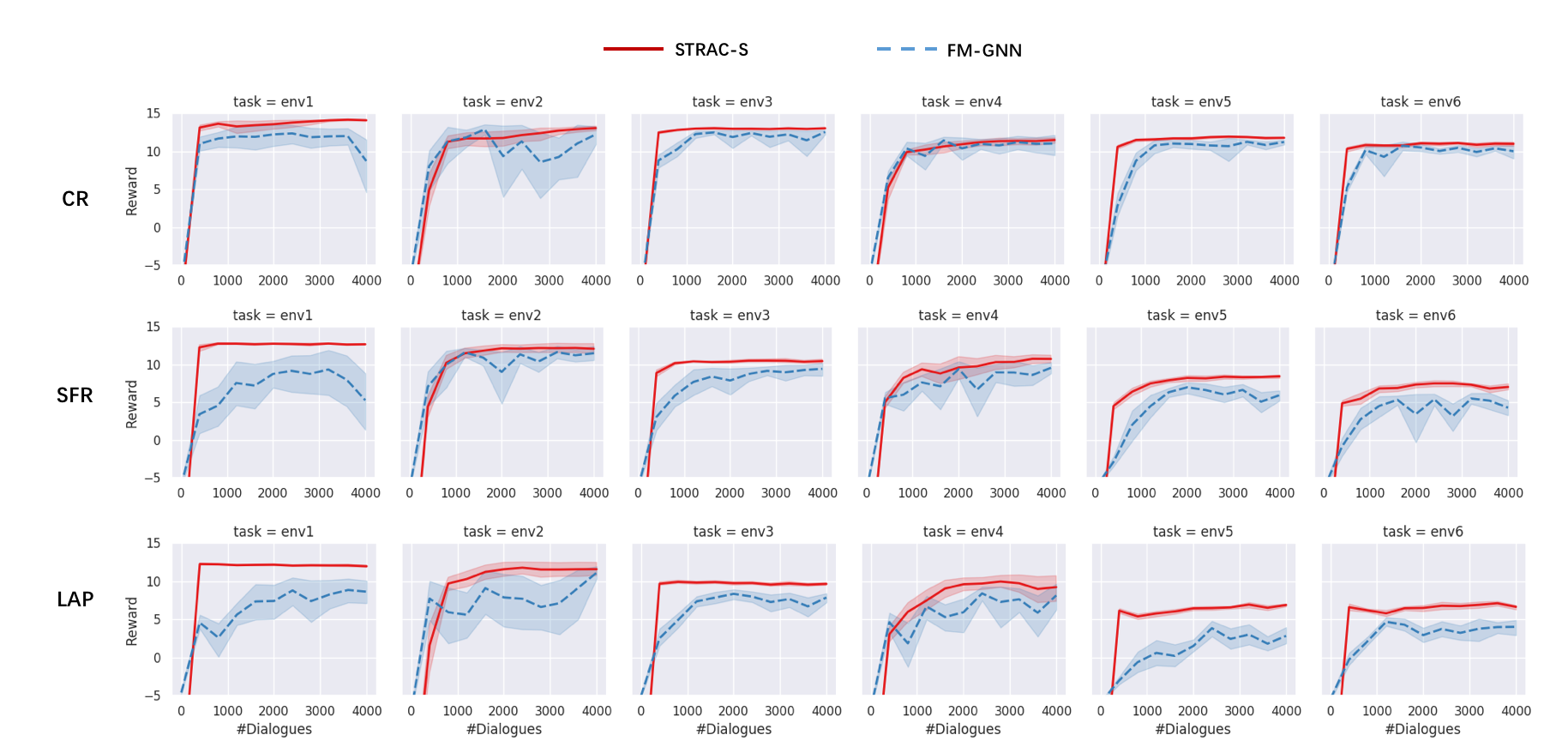}
\includegraphics[width=\textwidth]{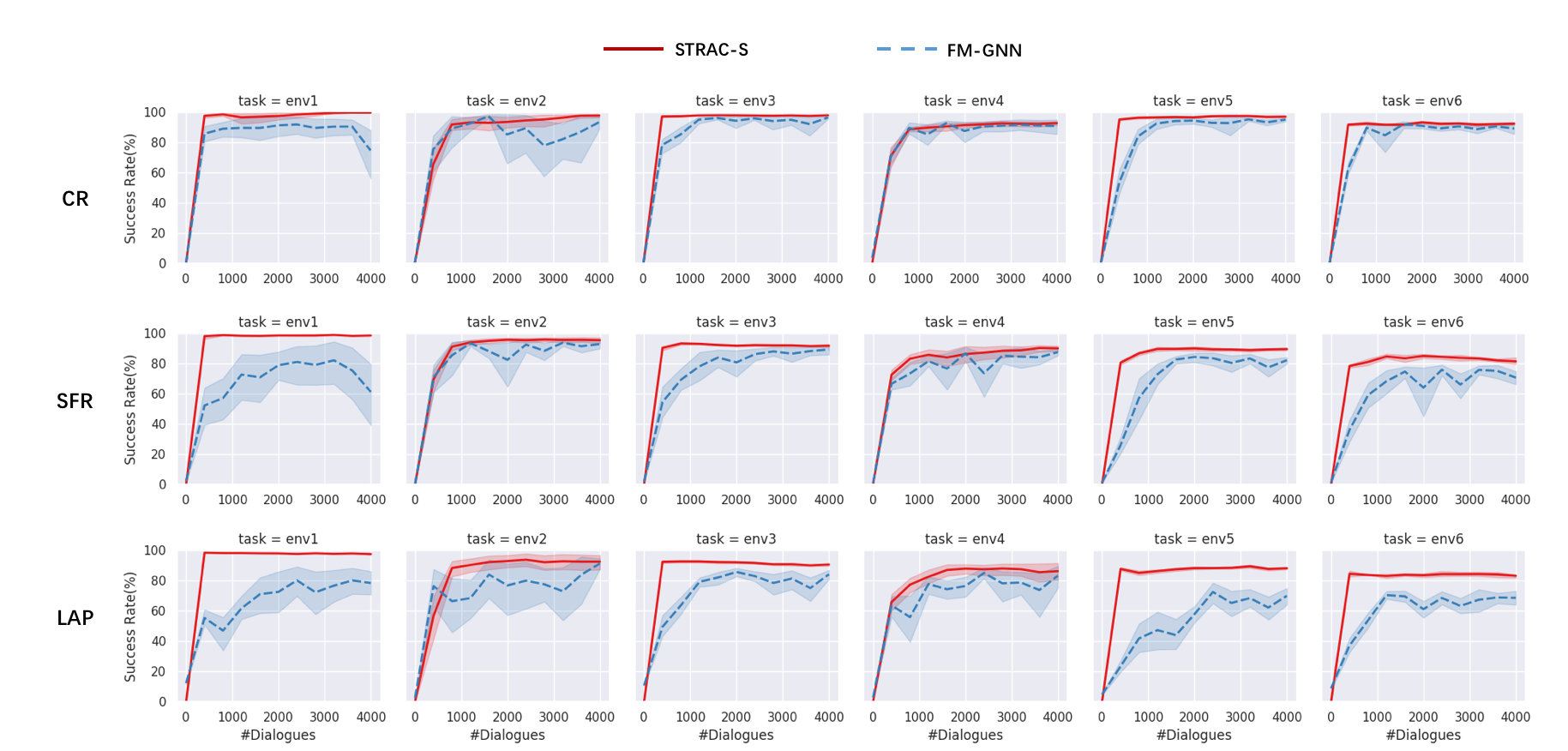}
\caption{The learning curves of reward and success rate for two GNN-based dialogue policy (STRAC-S and FM-GNN) on 18 different tasks.}
\label{fig:aclexp1}
\end{figure*}

\begin{figure*}[htbp!]
\centering
\includegraphics[width=\textwidth]{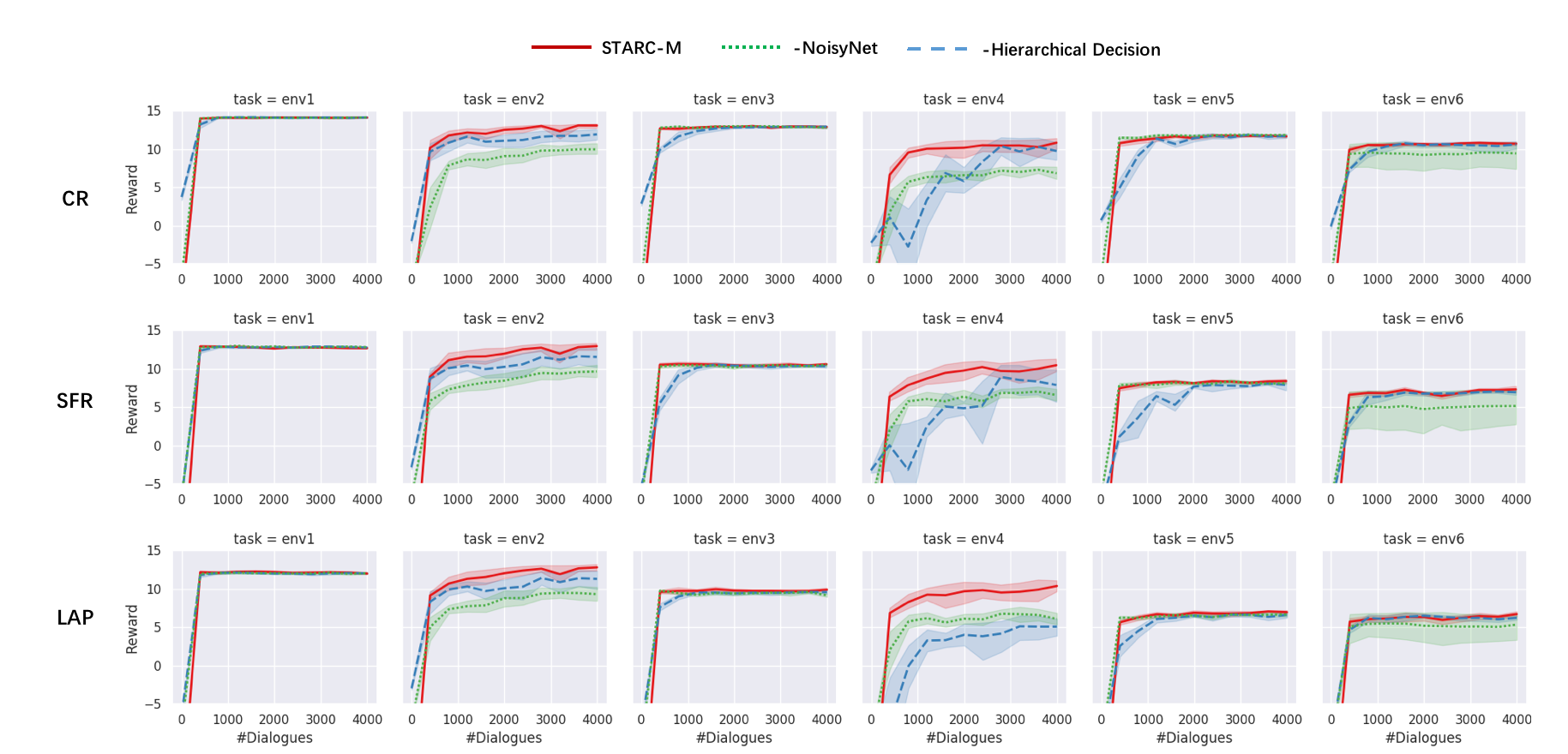}
\caption{The learning curves of reward for two ablation experiments without hierarchical decision-making mechanism or NoisyNet. }
\label{fig:aclexp3}
\end{figure*} 

\begin{figure}[htbp!]
\centering
\includegraphics[width=0.47\textwidth]{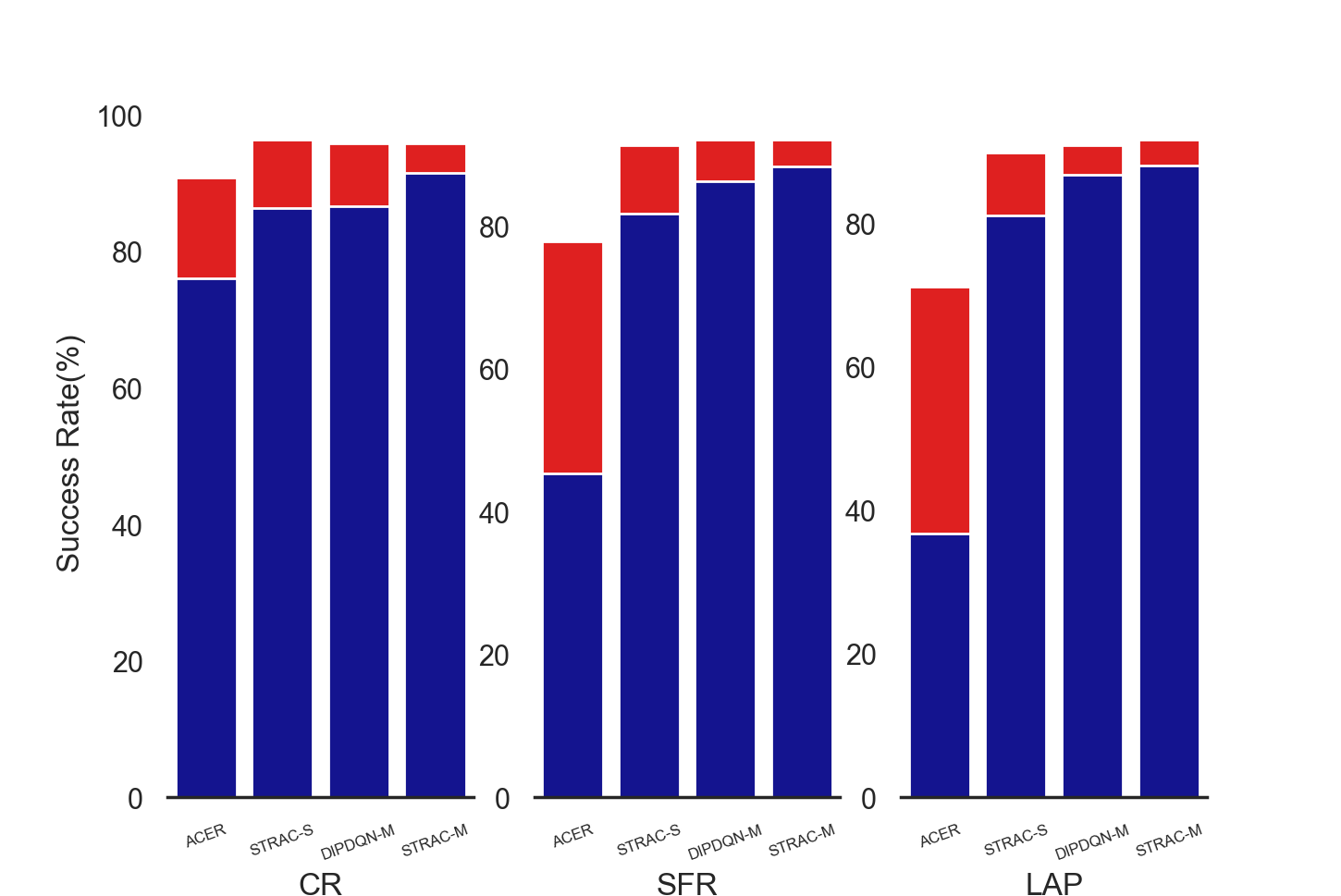}
\caption{The {\color{blue} blue bar} represents the average success rate of six different environments at 400 trajectories and the {\color{red} red bar} represents the average success rate at 4000 trajectories.  }
\label{fig:hist}
\end{figure}

For policy function $\pi_{\theta}$, the parameters $\theta$ are updated according to the policy gradient described in Equation \ref{eq:vgradient}. In order to encourage exploration, we also add an entropy bonus in the object function. Thus the total gradient for updating $\theta$ is as follows: 
\begin{align}
\nabla_{\theta} = &\lambda_{1}\rho_k \nabla_{\theta} \log \pi_{\theta}(a_k|\mathbf{b}_k)(r_k+\gamma v_{k+1} - V_{\beta}(\mathbf{b}_k))  \notag \\
&- \lambda_{2} \nabla_{\theta} \sum_a \pi_{\theta}(a|\mathbf{b}_k)\log \pi_{\theta}(a|\mathbf{b}_k),
\label{eq:ploss}
\end{align}
where $v_{k+1}$ is V-trace target at $\mathbf{b}_{k+1}$, and $\lambda_{1}$ and $\lambda_{2}$ are appropriate coefficients, which are hyper-parameters of the algorithm.

For the multi-task problem, the trajectories of experience are collected from different dialogue tasks. When training, we first sample a minibatch from each task respectively and then calculate the gradients of the state value function $V_{\beta}(\cdot)$ and the policy function $\pi_{\theta}(\cdot)$ on each minibatch according to Equation~\ref{eq:vloss} and Equation~\ref{eq:ploss}. When updating, we add these gradients together and update the parameters. The specific training procedure for the multi-task problem is shown in Algorithm~\ref{alg:strac}. For STRAC, a single task problem can be regarded as a special case of the multi-task problem, which has only one dialogue task.

\section{Experiments}
\label{sect:experiments}
In this section, we evaluate the performance of our proposed STRAC methods. First, we compare STRAC (STRAC-S) with another GNN-based dialogue policy (FM-GNN)~\cite{chen2018structured} and some popular policy optimization methods in single dialogue tasks. Then, we compare STRAC policy (STRAC-M) trained on the multiple tasks with STRAC-S and another generic dialogue policy DQNDIP-M, which is also trained parallel on the multiple tasks. We evaluate the performance of our method in the multi-task situation with limited dialogue data and sufficient dialogue data. Lastly, we design an ablation experiment to analyze the affected factors.

\subsection{Setup}
The PyDial benchmark has been used to deploy and evaluate dialogue policy models. It provides a set of 18 tasks (refer to~\cite{casanueva2017benchmarking} for a detailed description) consisting of 3 dialogue domains and 6 environments with different Semantic Error Rates (SER), a different configuration of action masks and user models (Standard or Unfriendly).

 Each S-agent has 3 actions 
 %(\emph{request}, \emph{confirm} and \emph{select})
 and I-agent has 5 actions. 
 %(\emph{inform by constraints, inform requested, inform alternatives, bye} and \emph{request more}). 
 More details  about actions as well as DIP features used here, please refer to \cite{casanueva2017benchmarking} and \cite{casanueva2018feudal}. 

The input module of the GNN policy is one hidden layer with size 40 and 250 for the S-agent and I-agent, respectively. There are two graph parsing layers. In the first graph parsing layer, the shapes of the message computation weights $\mathbf{W}_{S2S}^0$, $\mathbf{W}_{S2I}^0$ and $\mathbf{W}_{I2S}^0$ are $40*20$, $40*100$ and $250*20$ and the shapes of the representation update weight $\mathbf{W}_{S}^0$ and $\mathbf{W}_{I}^0$ are $20*40$ and $100*250$. The weight shapes of the second graph parsing layer are the same as the shapes of the corresponding weights. The outputs are one linear hidden layer, where the nodes with the same node type share the same output module. In order to drive the exploration, all the hidden layers in the neural network are noisy linear layers (NoisyNet)~\cite{fortunato2017noisy}. The activation function for all layers is a rectified linear unit (ReLU)~\cite{nair2010rectified}. 

The hyperparameters of STRAC are: $\overline{c}=5$, $\overline{\rho}=1$, $\gamma=0.99$, $n=5$, $\lambda_1=0.3$, $\lambda_2=0.001$. The learning rate $\alpha=0.0001$ and we use the Adam~\cite{kingma2014adam} optimizer. The size of a minibatch is 64. 

\textbf{Evaluation Setup:} When the dialogue does not terminate, at each turn of the dialogue, the reward is -1 to encourage a more efficient dialogue strategy. When the dialogue is terminated, if successful, the reward is 20. Otherwise, the reward is 0. There are two metrics to evaluate the performance of the dialogue policy, success rate and reward.

\subsection{Results of In-domain Policy}

In this subsection, we evaluate our proposed STRAC in single dialogue tasks (named \textbf{STRAC-S}). Here, we train models with 4000 dialogues or iterations. The total number of the training dialogues is broken down into milestones (20 milestones of 200 iterations each). At each milestone, there are 500 dialogues to test the performance of the dialogue policy. For each task, every model is trained with ten different random seeds (0 $\sim$ 9). The reward and success rate learning curves of STRAC-S, as well as FM-GNN, are shown in Fig.~\ref{fig:aclexp1}. FM-GNN is another kind of GNN-based dialogue policy, which combined the GNN-based policy model and Q-learning policy optimization method and has achieved state-of-the-art performance in previous literature. 

In the simple CR dialogue domain, which has fewer slots than the other two domains, all the methods can obtain considerable performance. 
In SFR and LAP, we can see that the more complex the dialogue task is, the more performance gain STRAC-S can achieve.
Additionally, we can further find that our method not only has better performance but also obtains more stable learning. 
Compared with FM-GNN in the initial training process, STRAC-S can converge stably to the final performance at a very fast speed. It demonstrates that the off-policy actor-critic approach can improve the speed of learning and stabilize the convergence procedure.

We have also implemented some popular dialogue policy methods, like GP-Sarsa~\cite{gavsic2015distributed}, DQN~\cite{chen-etal-2017-agent}, ACER~\cite{weisz2018sample} and FedualDQN, which are trained on single dialogue tasks. Their rewards and success rates are shown in Table~\ref{tab:results}. ACER is an improved version of the actor-critic approach. Compared with ACER, STRAC-S can achieve hierarchical decision-making. The decision-making process of FedualDQN is also hierarchical, but FedualDQN is a kind of Q-learning method and the hierarchical mechanism is not differentiable. Compared with these traditional approaches, STRAC-S can achieve the best average performance of six different environments on three dialogue domains, no matter when the training data is limited (400 dialogues at the second milestone) and sufficient (4000 dialogue).
It demonstrates that the structured policy model and advance off-policy training method in STRAC-S can efficiently improve the learning ability and learning speed.

\subsection{Results of Generic Policy}

In this subsection, we compare our proposed STRAC trained across multiple domains (\textbf{STRAC-M}) with ACER and STRAC-S. We further compare \textbf{STRAC-M} with the other state-of-the-art dialogue policy DQNDIP, which is trained on multiple tasks (named DQNDIP-M). We run all the experiments with limited dialogue data 400 iterations and sufficient dialogue data 4000 iterations. In STRAC-M and DQNDIP-M experiments, there is a generic policy with a set of shared parameters across three dialogue domains: Cambridge Restaurants (CR), San Francisco Restaurants (SFR) and Laptops (LAP). The optimization procedure of the unified dialogue agent is shown in Fig.~\ref{fig:LS}. When updating the share parameters, in a minibatch, there are three different kinds of dialogue experiences sampled from the replay memory. We use the dialogues from the same dialogue task to calculate the gradients of the shared parameters, respectively, and then add them together to update the shared parameters. We run STRAC-M experiments with 4000 iterations and the configurations are the same as STRAC-S experiments. 

% The rewards and success rates of STRAC-M and other baseline methods with 400 and 4000 iterations are shown in Table~\ref{tab:results}.

First, we compare the performance of STRAC-M with the performance of ACER, STRAC-S and DQNDIP-M. Their average performance of six different dialogue environments in CR domain, SFR domain and LAP domain is shown in Fig~\ref{fig:hist}. Compare STRAC-M with STRAC-S when the training iterations are limited (with only 400 dialogue data), we can find that STRAC-M learns much faster than STRAC-S. It demonstrates that the generic policy STRAC-M trained on all available data is very efficient. We can further see that STRAC-M achieves near-convergence performance after 400 iterations. 

Different from STRAC-M, DQNDIP-M is another generic dialogue policy, which directly combines the slot-dependent agent and slot-independent agent together. This combined agent is shared for all the slot-related agents (slot-independent agent is regarded as null slot  agent\cite{papangelis2019single}.). In other words, DQNDIP-M only considers the relations between the slot-dependent agent and the slot-independent agent by combining them together. Thus, DQNDIP-M ignores the relations among slot-dependent agents. Instead, our proposed STRAC uses a graph to model the relations among all the sub-agents. We can see that the average performance of STRAC-M is much better than that of the other approaches in Fig~\ref{fig:hist}, even than DQNDIP-M after 400 iterations. 
It demonstrates that STRAC-M is a good way to solve the cold start problem (the gap between bad initial performance and high-quality user experience) in real-world dialogue systems. 
% ACER is another off-policy actor-critic dialogue policy, which tries to improve sample efficiency. Compared with our structured actor-critic policy, ACER performs much worse in complex dialogue domains, like SFR and LAP.

As shown in Fig~\ref{fig:hist}, we can also find that the performance of STRAC-M is comparable and even better than the performance of STRAC-S, when the training iterations are sufficient. It demonstrates that the value scale problem, as discussed in Section~\ref{sect:background} impacts little on our proposed STRAC and STRAC has excellent transferability among different dialogue tasks. In Table~\ref{tab:results}, STRAC achieves new state-of-the-art performance on 13 (18 in total) tasks, whether the training iteration is limited or sufficient.

\subsection{Ablation Experiments}

In this subsection, we test two factors (hierarchical decision and NoisyNet) affecting the performance of STRAC. In the experiments, we take these two factors away, respectively, to test the effects. These two ablation experiments are introduced in detail below:
\begin{itemize}
%\item{\textbf{-Communication Mechanism}} In this experiment, there are not any communication parameters in the dialogue policy introduced in Section~\ref{sec:policy}.
\item{\textbf{-Hierarchical Decision}} The hierarchical operation $\mathbf{f}_{i}$ in Equation~\ref{eq:pref} is removed and replaced directly by $\mathbf{l}_{i}$.
\item{\textbf{-NoisyNet}} In this experiment, the noisy linear fully-connected layers are replaced by a normal linear fully-connected layer.
\end{itemize}
The learning curves of the ablation experiments in CR, SFR and LAP tasks are shown in Fig.~\ref{fig:aclexp3}. 

Without the hierarchical decision, we can find that the policy has obvious performance degradation in Env.2 and Env.4 without the action mask mechanism. It is worth noting that the dialogue space in Env.2 and Env.4 is larger than the other dialogue environments. It indicates that hierarchical decision is important for improving the learning ability and helpful for complex dialogue tasks, whose dialogue action space is large. In Fig.~\ref{fig:aclexp3}, we can find that the hierarchical decision mechanism gets more gain in LAP dialogue domain, which is more complicated than CR domain and SFR domain. Without NoisyNet, we can find that the variances increase obviously in Env.2, Env.4 and Env.6. It demonstrates that the exploration of NoisyNet is essential to stabilize the learning procedure.

\section{Conclusion}
\label{sect:conclusion}
This paper proposed a scalable distributed dialogue policy STRAC to train a generic dialogue policy on all available data collected from different dialogue tasks. STRAC increased scalability, stability and efficiency of the NN-based policy through combining structured dialogue policy and effective off-policy actor-critic algorithm. Compared with the traditional approaches, STRAC-M can be trained parallel on multiple tasks and gets the better performance, especially in the data-limited situation. Compared with another GNN-based policy optimization approach FM-GNN, the training process of STRAC is more stable and efficient. The final gains are more considerable in more complex environments. Compared with recent proposed generic policy DQNDIP-M, STRAC-M not only can be trained using the data from all the available dialogue tasks but also can model the relations among all the sub-agents. In future work, we will test STRAC with real users instead of the agenda-based user simulator~\cite{schatzmann2007agenda}.

\bibliographystyle{IEEEtran}
\bibliography{IEEEtran.bib}

\begin{IEEEbiography}[{\includegraphics[width=1in,height=1.25in,clip,keepaspectratio]{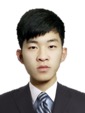}}]{Zhi Chen}
received his B.Eng. degree from the School of Software and Microelectronics, Northwestern Polytechnical University, in 2017. 
He is currently a Ph.D. student in the SpeechLab, Department of Computer Science and Engineering, Shanghai Jiao Tong University. His research interests include dialogue systems, reinforcement learning, and structured  deep learning. 
\end{IEEEbiography}

\begin{IEEEbiography}[{\includegraphics[width=1in,height=1.25in,clip,keepaspectratio]{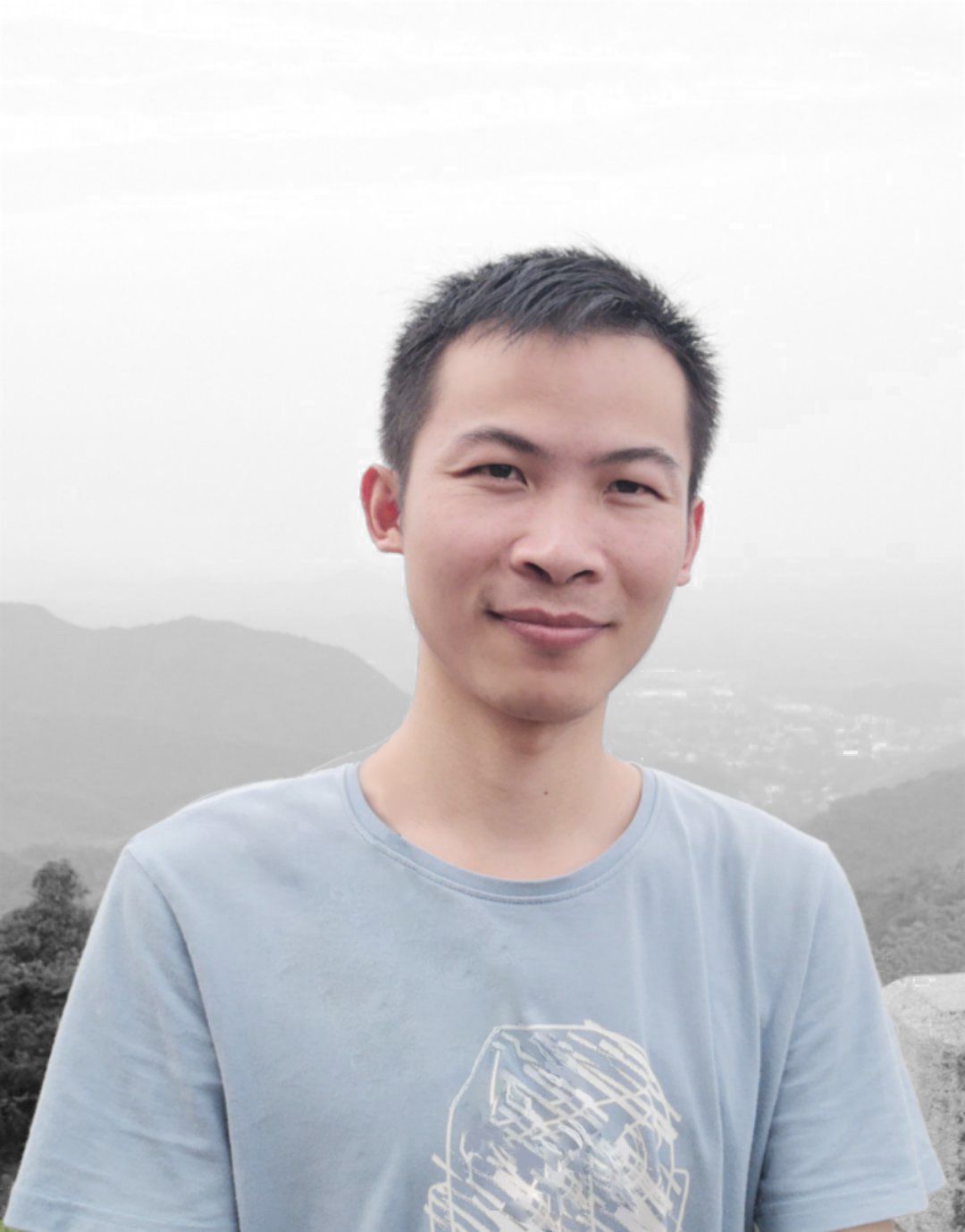}}]{Lu Chen}
received his Ph.D. degree in computer science from Shanghai Jiao Tong University (SJTU), China in 2020. Before coming to SJTU, he received a B.Eng. degree in computer science and engineering from Huazhong University of Science \& Technology (HUST), China in 2013. His research interests include dialogue systems, reinforcement learning, structured deep learning. He has authored/co-authored more than 20 journal articles (e.g. IEEE/ACM transactions) and peer-reviewed conference papers (e.g. ACL, EMNLP, AAAI).
\end{IEEEbiography}

\begin{IEEEbiography}[{\includegraphics[width=1in,height=1.25in,clip,keepaspectratio]{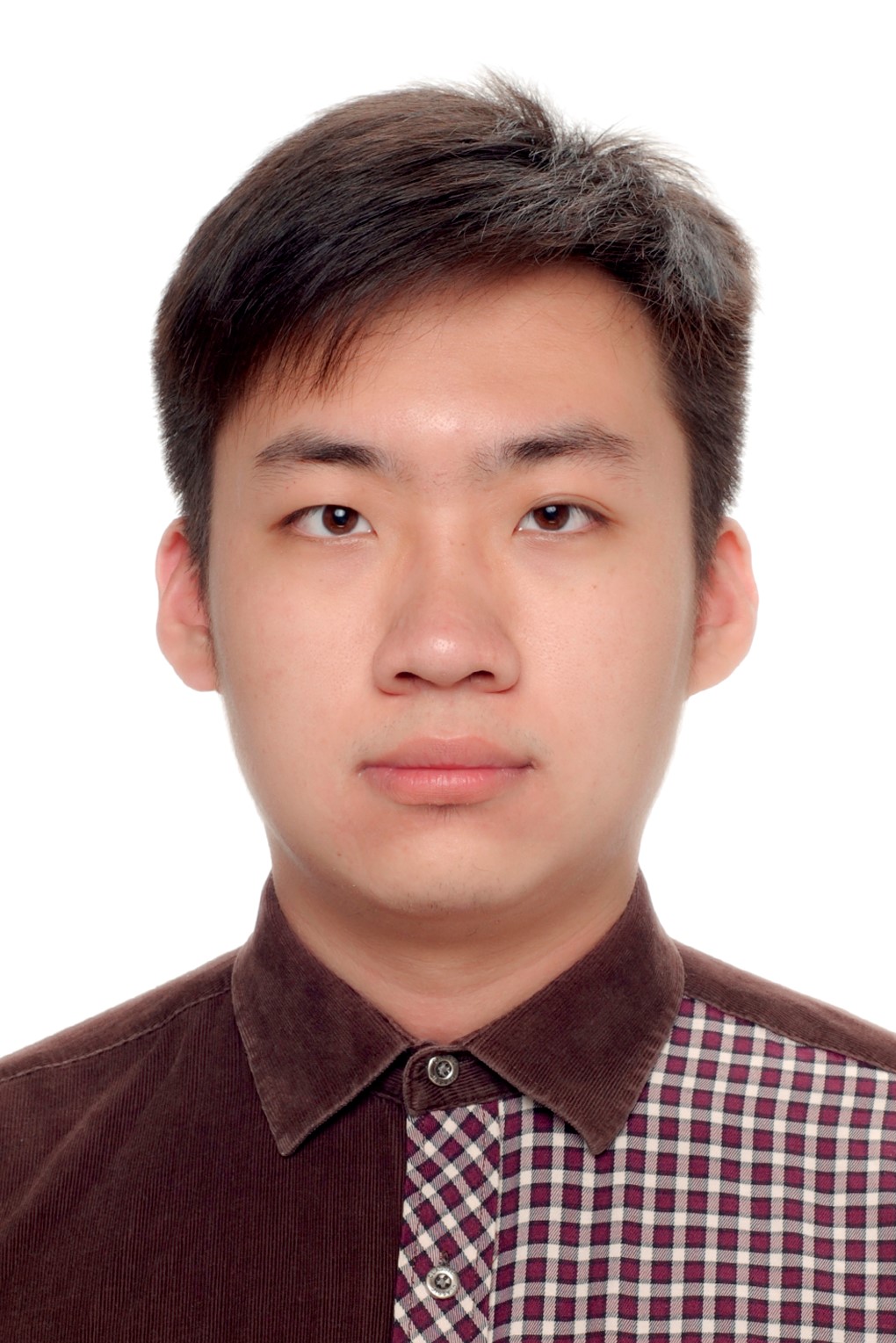}}]{Xiaoyuan Liu}
 received his B.Eng. degree from Zhiyuan College, Shanghai Jiao Tong University. From 2018 to 2020, he worked as an intern at SpeechLab, Department of Computer Science and Engineering, Shanghai Jiao Tong University. His research interests include natural language processing, computer security and systems.
\end{IEEEbiography}

\begin{IEEEbiography}[{\includegraphics[width=1in,height=1.25in,clip,keepaspectratio]{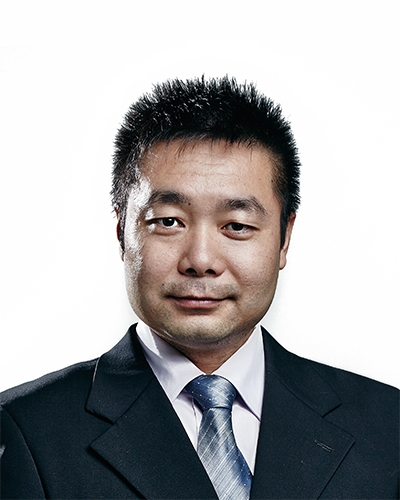}}]{Kai Yu}
 is a professor at Computer Science and Engineering Department, Shanghai Jiao Tong University, China. He received his B.Eng. and M.Sc. from Tsinghua University, China in 1999 and 2002, respectively. He then joined the Machine Intelligence Lab at the Engineering Department at Cambridge University, U.K., where he obtained his Ph.D. degree in 2006. His main research interests lie in the area of speech-based human machine interaction including speech recognition, synthesis, language understanding and dialogue management. He is a member of the IEEE Speech and Language Processing Technical Committee.
\end{IEEEbiography}

% or


% \appendices

% % you can choose not to have a title for an appendix
% % if you want by leaving the argument blank
% \section{}

% % use section* for acknowledgment
% \section*{Acknowledgment}


\ifCLASSOPTIONcaptionsoff
  \newpage
\fi

\end{document}